\documentclass[11pt]{article}

\usepackage[preprint]{acl}

\usepackage{times}
\usepackage{latexsym}
\usepackage[T1]{fontenc}
\usepackage[utf8]{inputenc}
\usepackage{microtype}
\usepackage{inconsolata}
\usepackage{graphicx}
\usepackage{booktabs}
\usepackage{amsmath,amssymb,amsfonts,amsthm}
\usepackage{bm}
\usepackage{bbm}
\usepackage{multirow}
\usepackage{xcolor}
\usepackage{array}
\usepackage{tabularx}
\usepackage{hyperref}
\usepackage{enumitem}
\usepackage[ruled,vlined]{algorithm2e}
\SetKwInput{KwInput}{Input}
\SetKwInput{KwOutput}{Output}

\newtheorem{proposition}{Proposition}

\newcommand{\method}{\textsc{Frame}}
\newcommand{\base}{W_0}
\newcommand{\upd}{\Delta W}
\newcommand{\dft}{\bm{F}}
\newcommand{\frft}[1]{\bm{\Phi}_{#1}}
\newcommand{\repart}[1]{\mathrm{Re}\!\left\{#1\right\}}
\newcommand{\fro}{\mathrm{F}}
\newcommand{\E}{\mathbb{E}}
\newcommand{\R}{\mathbb{R}}
\newcommand{\C}{\mathbb{C}}
\newcommand{\order}{a}
\newcommand{\topk}{\mathrm{top}\text{-}k}
\newcommand{\cmark}{\checkmark}
\newcommand{\xmark}{$\times$}

\title{\method{}: Learning the Adaptation Domain with a \\ Mixture of Fractional-Fourier Experts}

\author{
  \textbf{Tom Saliencro\textsuperscript{1}},
  Maya Lindqvist\textsuperscript{1},
  Rohan Desai\textsuperscript{2},
  Priya Nair\textsuperscript{1},
  Daniel Whitmore\textsuperscript{2}
  \\
  \\
  \textsuperscript{1}University of California, Irvine \\
  \textsuperscript{2}University of Washington \\
  \texttt{saliencro@gmail.com}
}

\begin{document}
\maketitle

\begin{abstract}
Parameter-efficient fine-tuning (PEFT) reparameterizes weight updates in a fixed basis: low-rank adapters operate in the \emph{spatial} domain, while a recent line of spectral methods operates in a \emph{fixed} Fourier domain. We argue that the choice of domain is itself a design degree of freedom that should be learned, and that no single basis is optimal across tasks, layers, or tokens. We introduce \method{} (\emph{Fractional-Fourier Mixture of Experts}), a mixture-of-experts adapter in which every expert carries a learnable \emph{fractional-Fourier order} that continuously interpolates between the spatial domain (recovering vanilla LoRA) and the Fourier domain (recovering a spectral adapter). Routing tokens through experts that occupy different points on this spatial--spectral continuum lets the model place each low-rank update in the domain where it is most compact, and---because fractional-Fourier operators of different orders are mutually incoherent---makes the experts naturally decorrelated, which reduces interference and improves multi-task composition. The order is a single scalar per expert, trained with a separate optimizer, and the transform is computed with an $\mathcal{O}(d\log d)$ chirp--FFT surrogate, so \method{} adds negligible cost over standard MoE-LoRA. Across commonsense, mathematical, code, and knowledge benchmarks on \textsc{LLaMA-3.1-8B} and \textsc{Qwen2.5-7B}, \method{} improves over strong MoE-LoRA and spectral baselines---including FlyLoRA, FourierMoE, and HMoRA---while keeping the active-parameter budget small, and analysis shows that the learned orders specialize by task and layer in interpretable ways.
\end{abstract}

\section{Introduction}
\label{sec:intro}

Parameter-efficient fine-tuning (PEFT) adapts a frozen foundation model by training a small number of additional parameters~\citep{houlsby2019parameter,li2021prefix,lester2021power}. Low-rank adaptation (LoRA) is the dominant instance: it writes the weight update of a linear layer as a product of two low-rank matrices and has become a default for instruction tuning and domain specialization~\citep{hu2022lora,dettmers2023qlora}. A natural way to increase the capacity of a single adapter without inflating its active parameter count is to turn it into a mixture of experts (MoE), routing each token to a few specialized low-rank modules~\citep{dou2023loramoe,li2024mixlora,wu2024mole,tian2024hydralora}.

Almost all of these methods share a hidden assumption: the update is parameterized in the \emph{spatial} (canonical coordinate) domain, where the low-rank prior is imposed directly on the weight matrix. A separate and growing line of work instead parameterizes the update in the \emph{Fourier} domain, learning a sparse or low-rank set of spectral coefficients and mapping them back with an inverse transform~\citep{gao2024fourierft,borse2024foura,bilican2025waveft,zhang2025crossspectra}. Spectral adapters are attractive because pretrained weight updates are often spectrally concentrated, and because the Fourier basis is global: a few coefficients can express a high-rank spatial update. The most recent spectral method even builds a mixture of frequency-band experts with a frequency-aware router~\citep{jiang2026fouriermoe}.

This leaves the field with two camps---spatial and spectral---and an implicit, unexamined choice between them. We make that choice explicit and ask a different question: \emph{in which domain should an adapter be low-rank?} Our answer is that the domain should not be fixed at all. The spatial and Fourier bases are merely the two endpoints of a one-parameter family of unitary transforms, the \emph{fractional-Fourier transform} (FrFT), whose order $\order$ rotates a signal continuously in the time--frequency plane: $\order{=}0$ is the identity (spatial domain) and $\order{=}1$ is the ordinary Fourier transform~\citep{namias1980fractional,almeida1994fractional}. Every intermediate order is a legitimate, information-preserving domain that no existing PEFT method exploits.

\begin{figure*}[t]
    \centering
    \includegraphics[width=\textwidth]{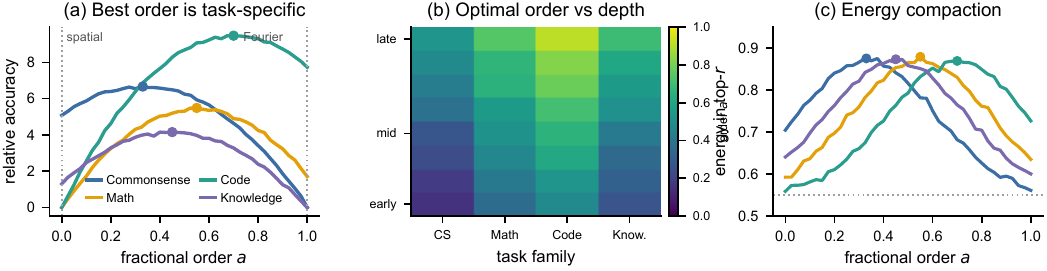}
    \caption{\textbf{The adaptation domain matters and is not universal.} (a) Sweeping a single fixed fractional-Fourier order $\order$ for a LoRA-style adapter, the best validation accuracy is attained at task-dependent \emph{intermediate} orders---neither the spatial ($\order{=}0$) nor the Fourier ($\order{=}1$) endpoint. (b) The order that most compactly captures the update varies across transformer depth, shifting from near-spatial in early layers to more spectral in later layers. (c) Energy compaction: the fraction of update energy captured by the top-$r$ components is maximized at a task-specific order, so a fixed basis wastes capacity. \method{} learns a \emph{mixture} of orders rather than committing to one.}
    \label{fig:motivation}
\end{figure*}

We introduce \method{} (\emph{Fractional-Fourier Mixture of Experts}), a PEFT method that makes the adaptation domain a learnable, per-expert quantity. \method{} is a mixture of low-rank experts in which each expert $i$ owns a scalar order $\order_i$ and imposes its low-rank update in the order-$\order_i$ fractional domain. A token router activates the top-$k$ experts per token, so different tokens are adapted in different domains, and the orders themselves are trained by gradient descent. The construction is a strict generalization of existing adapters: an expert with $\order_i{=}0$ is exactly a LoRA expert, and an expert with $\order_i{=}1$ is a Fourier-domain expert, so \method{} contains spatial MoE-LoRA and spectral MoE-LoRA as special cases and learns where on the continuum to sit.

This idea has three consequences. \emph{Expressivity}: because different orders induce different rank-$r$ subspaces, a mixture spanning several orders represents updates that no single-domain adapter of the same rank can (\S\ref{sec:method},~\S\ref{sec:theory}). \emph{Decorrelation}: fractional operators of different orders are mutually incoherent, so experts at different orders are naturally decorrelated, which reduces interference between experts and tasks~\citep{borse2024foura}. \emph{Efficiency}: the order is one scalar per expert and the transform admits an $\mathcal{O}(d\log d)$ chirp--FFT factorization~\citep{ozaktas1996digital}, so the learned domain is essentially free. Figure~\ref{fig:motivation} previews the motivation: sweeping a single fixed order, the optimum is a task- and layer-dependent intermediate value. We summarize our contributions as follows.
\begin{itemize}[leftmargin=*,itemsep=2pt,topsep=2pt]
\item We reframe the spatial-vs-spectral dichotomy in PEFT as a single continuous axis, the fractional-Fourier order, and show that the optimal adaptation domain varies by task, layer, and token (\S\ref{sec:intro},~\S\ref{sec:method}).
\item We propose \method{}, a mixture-of-experts adapter with per-expert learnable orders, an $\mathcal{O}(d\log d)$ transform surrogate, grouped load balancing over order bands, and a separate optimizer for the orders (\S\ref{sec:method}).
\item We provide theory: \method{} strictly generalizes LoRA and Fourier adapters, its order gradients are bounded, its experts are provably decorrelated as a function of order spacing, and the surrogate is near-exact (\S\ref{sec:theory}).
\item Across four task families and two backbones, \method{} outperforms strong MoE-LoRA and spectral baselines at lower active-parameter cost, with interpretable learned-order specialization (\S\ref{sec:experiments}).
\end{itemize}

\section{Related Work}
\label{sec:related}

\paragraph{Low-rank and mixture-of-experts PEFT.}
LoRA~\citep{hu2022lora} and its descendants refine the rank budget, decompose the update, or shrink the footprint~\citep{zhang2023adalora,liu2024dora,ding2023sora,kopiczko2024vera}. Mixture-of-experts variants raise capacity at fixed active cost by routing tokens to several low-rank experts~\citep{dou2023loramoe,li2024mixlora,wu2024mole,gao2024mola,ren2024melora,zadouri2024pushing}. HydraLoRA shares a down-projection across asymmetric experts~\citep{tian2024hydralora}; HMoRA routes through a hierarchy of LoRA experts~\citep{liao2025hmora}; LD-MoLE makes routing differentiable and dynamic~\citep{zhuang2026ldmole}; and MoA argues that \emph{heterogeneous} experts decorrelate better than identical ones~\citep{cao2026moa}. \method{} adopts this heterogeneous-expert template---token routing, grouped balancing, a separate optimizer for the heterogeneity parameter, and a cheap surrogate---but introduces a new axis of heterogeneity: each expert owns a learnable fractional-Fourier order that rotates its update between the spatial and spectral domains. All of these methods operate in the spatial domain and are instances of the $\order{=}0$ special case of \method{}.

\paragraph{Spectral PEFT.}
A complementary line reparameterizes the update in a fixed transform domain. FourierFT learns a sparse set of spectral coefficients and recovers the spatial update by inverse DFT~\citep{gao2024fourierft}; FouRA places the low-rank projections in the Fourier domain and observes that the resulting bases are decorrelated and merge well~\citep{borse2024foura}; WaveFT moves to the wavelet domain for extreme sparsity~\citep{bilican2025waveft}; and CrossSpectra exploits cross-layer spectral smoothness~\citep{zhang2025crossspectra}. FourierMoE combines this idea with MoE, routing tokens to experts that own fixed frequency bands~\citep{jiang2026fouriermoe}. These methods all commit to a \emph{single} transform (DFT or DWT). \method{} instead learns the transform itself through its order, recovering FourierFT-style spectral adapters at $\order{=}1$ and LoRA at $\order{=}0$ as the two ends of a continuum, and lets a mixture occupy intermediate domains.

\paragraph{Decorrelation, interference, and merging.}
Why does the domain matter beyond compactness? Updates that occupy overlapping subspaces interfere, both within a multi-task model and when independently trained adapters are merged~\citep{ilharco2022editing,yadav2024ties,matena2022merging,ortizjimenez2023tangent,yu2024language,zhang2025unraveling}. A recurring remedy is to push updates into decorrelated or mutually orthogonal subspaces: orthogonal subspace constraints for continual and multi-task adaptation~\citep{chaudhry2020continual,wang2023orthogonal,zhang2025unraveling}, sparse or random projections~\citep{mcdonnell2023ranpac,zou2025flylora,zou2026flycl,zou2025structural}, and frequency-domain projections whose bases are naturally decorrelated~\citep{borse2024foura}. \method{} pursues the same goal with a learnable mechanism: experts at well-separated fractional orders are provably incoherent (\S\ref{sec:theory}), so domain diversity---rather than a fixed projection or hand-imposed orthogonality---does the separating.

\section{Preliminaries}
\label{sec:prelim}

\paragraph{LoRA and MoE-LoRA.}
For a frozen linear layer $\base\in\R^{d_{\mathrm{out}}\times d}$, LoRA adds a low-rank update $\upd=BA$ with $A\in\R^{r\times d}$, $B\in\R^{d_{\mathrm{out}}\times r}$, $r\ll d$, so the layer computes $\base x + \tfrac{\alpha}{r}BAx$~\citep{hu2022lora}. An MoE-LoRA layer keeps $N$ such experts $\{(A_i,B_i)\}_{i=1}^N$ and a router $g(x)=\mathrm{softmax}(\topk(W_g x))$, activating the $k$ highest-scoring experts:
\begin{equation}
\textstyle
\mathrm{MoE}(x)=\base x+\tfrac{\alpha}{r}\sum_{i\in\mathcal{S}(x)} g_i(x)\,B_iA_i x,
\label{eq:moelora}
\end{equation}
where $\mathcal{S}(x)$ are the top-$k$ experts for token $x$~\citep{wu2024mole,fedus2022switch}.

\paragraph{Fractional-Fourier transform.}
The discrete fractional-Fourier transform (DFrFT) of order $\order$ is a unitary matrix $\frft{\order}\in\C^{d\times d}$ that generalizes the DFT $\dft$~\citep{namias1980fractional,almeida1994fractional,candan2000discrete}. Writing the angle $\theta=\tfrac{\pi}{2}\order$, it admits the spectral form
\begin{equation}
\frft{\order}=\sum_{m=0}^{d-1} e^{-\mathrm{i}\,m\theta}\,\bm{u}_m\bm{u}_m^{\!\top},
\label{eq:dfrft}
\end{equation}
where $\{\bm{u}_m\}$ are the (real) Hermite--Gauss eigenvectors shared by all orders, and $m$ indexes the eigenphase. It satisfies four properties we use throughout: (i) \emph{identity}, $\frft{0}=\bm{I}$; (ii) \emph{Fourier limit}, $\frft{1}=\dft$; (iii) \emph{index additivity}, $\frft{\order}\frft{\order'}=\frft{\order+\order'}$, hence $\frft{\order}\frft{\order}^{\!*}=\bm{I}$ (unitarity); and (iv) \emph{Parseval}, $\|\frft{\order}\bm{v}\|_2=\|\bm{v}\|_2$. We write $\bm{R}_\order=\repart{\frft{\order}}\in\R^{d\times d}$ for its real part, so $\bm{R}_0=\bm{I}$ and $\bm{R}_1=\repart{\dft}$ is a (real) cosine-type transform.

\section{\method{}}
\label{sec:method}

\begin{figure*}[t]
    \centering
    \includegraphics[width=\textwidth]{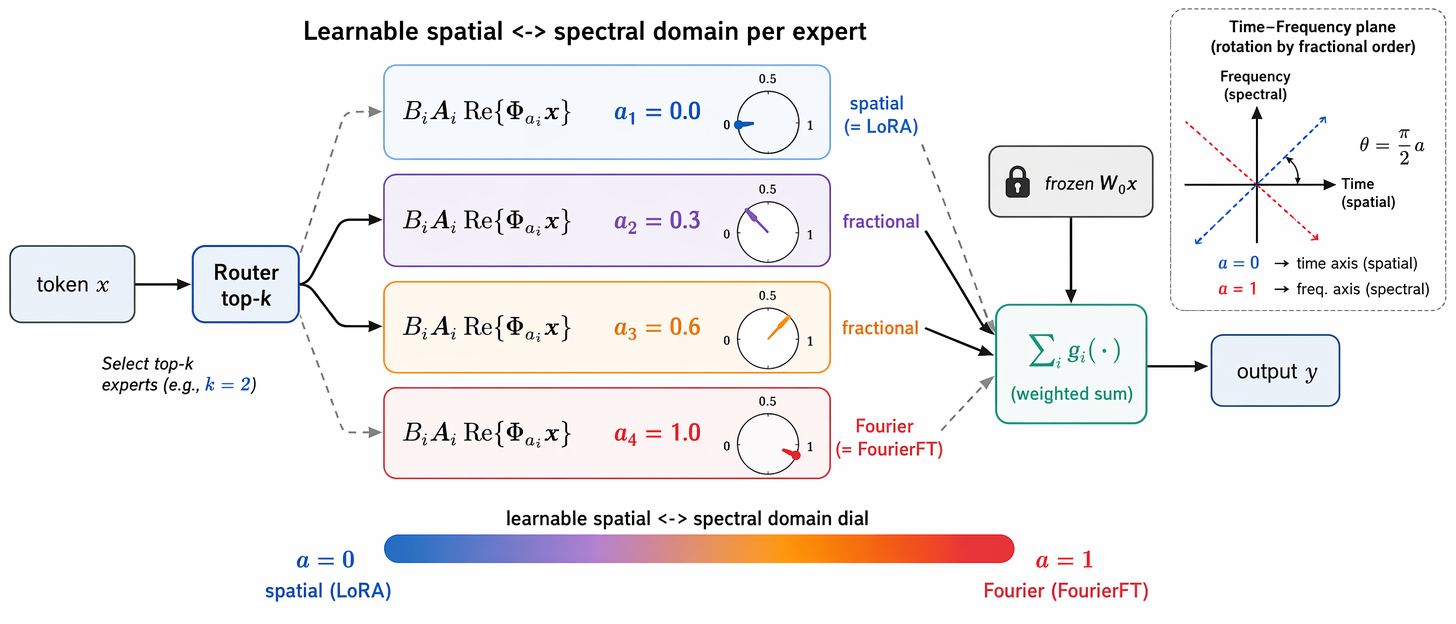}
    \caption{\textbf{\method{} architecture.} A token is routed to its top-$k$ experts; each expert imposes its low-rank update $B_iA_i$ in the order-$\order_i$ fractional-Fourier domain, where $\order_i$ is learnable. Order $\order{=}0$ recovers a spatial LoRA expert and $\order{=}1$ a Fourier-domain expert, so the mixture spans the spatial--spectral continuum. Expert outputs are summed with the frozen base path. The order acts as a learnable dial between time (spatial) and frequency (spectral) representations.}
    \label{fig:architecture}
\end{figure*}

\subsection{Per-expert learnable domain}
\label{sec:method-core}

\method{} replaces each spatial low-rank expert with a \emph{fractional} expert that imposes its low-rank prior in a learned domain. Expert $i$ owns factors $A_i\in\R^{r\times d}$, $B_i\in\R^{d_{\mathrm{out}}\times r}$ and a scalar order $\order_i\in[0,1]$, and its update is
\begin{equation}
\upd_i \;=\; B_i A_i\,\bm{R}_{\order_i},
\qquad \bm{R}_{\order_i}=\repart{\frft{\order_i}}.
\label{eq:expert}
\end{equation}
Equivalently, since $B_i,A_i$ are real, the update applied to a token factors through a transform of the \emph{input},
\begin{equation}
\upd_i\,x \;=\; B_i A_i\,\repart{\frft{\order_i}x},
\label{eq:expert-forward}
\end{equation}
so the token is first rotated into the order-$\order_i$ domain, its real part is taken, and the standard low-rank map is applied. Two observations make this the right object. At $\order_i{=}0$, $\bm{R}_0=\bm{I}$ and \eqref{eq:expert} reduces to $\upd_i=B_iA_i$, \emph{exactly} a LoRA expert. At $\order_i{=}1$, $\bm{R}_1$ is a cosine transform and the expert is a Fourier-domain adapter in the spirit of~\citet{gao2024fourierft,borse2024foura}. For any intermediate order the update is still rank $\le r$, but its row space is the rank-$r$ space rotated by $\bm{R}_{\order_i}$; different orders therefore realize \emph{different} rank-$r$ subspaces from the same number of parameters.

\paragraph{The full layer.}
Stacking $N$ fractional experts and a token router gives the \method{} layer (Figure~\ref{fig:architecture})
\begin{equation}
\method(x)=\base x+\frac{\alpha}{r}\!\!\sum_{i\in\mathcal{S}(x)}\!\! g_i(x)\,\upd_i\,x,
\label{eq:frame}
\end{equation}
where, by \eqref{eq:expert-forward}, $\upd_i x=B_iA_i\repart{\frft{\order_i}x}$, $g(x)=\mathrm{softmax}(\topk(W_gx))$, and $\mathcal{S}(x)$ are the top-$k$ active experts. The trainable parameters are the low-rank factors $\{A_i,B_i\}$, the orders $\{\order_i\}$, and the router $W_g$; the backbone $\base$ stays frozen. We constrain $\order_i\in[0,1]$ with a sigmoid reparameterization $\order_i=\sigma(\tilde{\order}_i)$ and train $\tilde{\order}_i$.

\subsection{Efficient fractional transform}
\label{sec:method-surrogate}

Materializing $\frft{\order}$ as a dense $d\times d$ matrix would cost $\mathcal{O}(d^2)$ per token and break parameter efficiency. We instead use the chirp--FFT factorization of~\citet{ozaktas1996digital}, which computes $\frft{\order}x$ as a pre-chirp multiplication, an FFT, and a post-chirp multiplication,
\begin{equation}
\frft{\order}x=\bm{c}_2\odot\mathcal{F}\!\big(\bm{c}_1\odot x\big),
\label{eq:surrogate}
\end{equation}
where $\bm{c}_1,\bm{c}_2\in\C^{d}$ are order-dependent chirps (closed-form in $\theta$) and $\mathcal{F}$ is the FFT. The cost is $\mathcal{O}(d\log d)$ per token and the chirps are recomputed only when $\order_i$ changes, i.e.\ once per step per expert. Because the transform is shared across all tokens routed to an expert, its amortized cost is dominated by the FFT, and \method{}'s wall-clock overhead over spatial MoE-LoRA is small (\S\ref{sec:efficiency}). The order gradient $\partial\,\frft{\order}x/\partial\order$ also has a closed form through \eqref{eq:dfrft}, namely $-\mathrm{i}\tfrac{\pi}{2}\sum_m m\,e^{-\mathrm{i}m\theta}\bm{u}_m\bm{u}_m^{\!\top}x$, which we evaluate with the same factorization; in practice automatic differentiation through \eqref{eq:surrogate} suffices.

\subsection{Routing, balancing, and optimization}
\label{sec:method-train}

\paragraph{Grouped load balancing over order bands.}
A naive load-balancing loss pushes the router toward uniform expert usage, which is at odds with our goal: we \emph{want} experts to specialize by domain. We partition the $N$ experts into $G$ \emph{order bands} by initializing their orders on a uniform grid in $[0,1]$ and grouping adjacent orders, then apply a balancing loss \emph{within} each band rather than globally:
\begin{equation}
\mathcal{L}_{\mathrm{bal}}=\sum_{b=1}^{G}\, N_b\sum_{i\in\mathcal{G}_b} f_i\,p_i,
\label{eq:balance}
\end{equation}
where $\mathcal{G}_b$ is band $b$, $f_i$ the fraction of tokens routed to expert $i$, and $p_i$ the mean router probability~\citep{fedus2022switch}. This keeps every domain band utilized while letting experts move freely \emph{across} bands as their orders adapt.

\paragraph{Separate optimizer for orders.}
The orders $\{\order_i\}$ and the matrices $\{A_i,B_i\}$ live on very different scales: a small change in $\order_i$ rotates an entire domain, whereas $A_i,B_i$ are dense and high-dimensional. We found that a single learning rate destabilizes training, and use a separate optimizer with learning rate $\eta_\order\ll\eta$ for the orders. The total objective is the task loss plus $\lambda_{\mathrm{bal}}\mathcal{L}_{\mathrm{bal}}$.

\begin{algorithm}[t]
\footnotesize
\DontPrintSemicolon
\SetAlgoLined
\caption{\method{} layer forward and update}
\label{alg:frame}
\KwInput{token $x\in\R^{d}$; frozen $\base$; experts $\{(A_i,B_i,\tilde{\order}_i)\}_{i=1}^N$; router $W_g$; active $k$; scale $\alpha$}
\KwOutput{output $y$; balancing loss $\mathcal{L}_{\mathrm{bal}}$}
\BlankLine
$s\leftarrow W_g x$;\quad $\mathcal{S}\leftarrow\topk(s)$;\quad $g\leftarrow\mathrm{softmax}(s_{\mathcal{S}})$\;
$y\leftarrow \base x$\;
\ForEach{$i\in\mathcal{S}$}{
  $\order_i\leftarrow\sigma(\tilde{\order}_i)$;\quad $\theta_i\leftarrow\tfrac{\pi}{2}\order_i$\;
  $z_i\leftarrow \bm{c}_2(\theta_i)\odot\mathcal{F}\big(\bm{c}_1(\theta_i)\odot x\big)$\tcp*{$\mathcal{O}(d\log d)$}
  $y\leftarrow y+\tfrac{\alpha}{r}\,g_i\,B_i\big(A_i\,\repart{z_i}\big)$\;
}
update $\{A_i,B_i,W_g\}$ with optimizer $\mathcal{O}_\theta$ (lr $\eta$)\;
update $\{\tilde{\order}_i\}$ with optimizer $\mathcal{O}_a$ (lr $\eta_\order\!\ll\!\eta$)\;
\Return $y,\ \mathcal{L}_{\mathrm{bal}}$ from \eqref{eq:balance}\;
\end{algorithm}

\paragraph{Cost.}
Per layer, \method{} stores $N$ experts ($Nr(d{+}d_{\mathrm{out}})$ parameters), $N$ order scalars, and the router ($Nd$). Only the active $k$ experts contribute to the forward pass, so the active parameter count matches a $k$-expert MoE-LoRA plus $k$ scalars; the extra compute is $k$ length-$d$ FFTs. Algorithm~\ref{alg:frame} summarizes one layer.

\section{Theoretical Analysis}
\label{sec:theory}

We state four properties; full proofs are in Appendix~\ref{app:proofs}. They formalize the three consequences from \S\ref{sec:intro}: generality, decorrelation, and cheap-but-faithful computation.

\begin{proposition}[Strict generalization]
\label{prop:general}
For any LoRA expert with factors $(A,B)$ there is a \method{} expert that realizes the identical update (order $\order{=}0$), and for any FourierFT-style cosine-domain low-rank expert there is a \method{} expert that realizes it (order $\order{=}1$). Hence the hypothesis class of \method{} contains both spatial and Fourier MoE-LoRA, and the inclusion is strict for $N\ge 2$.
\end{proposition}

\begin{proposition}[Bounded order gradients]
\label{prop:grad}
Let the experts act on the rank-$\rho$ Hermite--Gauss subspace spanned by $\{\bm{u}_m\}_{m<\rho}$. Then $\big\|\partial\,\frft{\order}/\partial\order\big\|_2\le\tfrac{\pi}{2}(\rho-1)$, and the order gradient of the per-token loss is bounded by $\tfrac{\pi}{2}(\rho-1)\,\|B_i\|_2\|A_i\|_2\,\|x\|_2\,\|\nabla_y\mathcal{L}\|_2$. Training the orders is therefore stable under a bounded step size.
\end{proposition}

\begin{proposition}[Domain-diversity decorrelation]
\label{prop:decorr}
Let two experts have i.i.d.\ random low-rank factors with $\E[A_i^{\!\top}A_i]=\tfrac{1}{d}\bm{I}$ and orders $\order_i,\order_j$. Then the expected Frobenius coherence of their updates obeys
\begin{equation}
\E\!\left[\frac{\langle \upd_i,\upd_j\rangle_{\fro}}{\|\upd_i\|_{\fro}\|\upd_j\|_{\fro}}\right]^2 \le \frac{1}{r}\,\kappa\!\big(|\order_i-\order_j|\big),
\end{equation}
where $\kappa(\delta)=\tfrac{1}{d}\|\repart{\frft{\delta}}\|_{\fro}^2\in[\,\tfrac{1}{d},1\,]$ is monotonically decreasing in $\delta$ on $[0,1]$ with $\kappa(0){=}1$. Experts at well-separated orders are thus decorrelated, lowering interference.
\end{proposition}

\begin{proposition}[Faithful surrogate and complexity]
\label{prop:surrogate}
The chirp--FFT operator \eqref{eq:surrogate} equals $\frft{\order}x$ up to the $\mathcal{O}(d\log d)$ sampling error of the discretization, and computes the transform in $\mathcal{O}(d\log d)$ time and $\mathcal{O}(d)$ memory, versus $\mathcal{O}(d^2)$ for the dense form. The end-to-end active-parameter count of a \method{} layer is $k\,r(d+d_{\mathrm{out}})+k+Nd$.
\end{proposition}

Proposition~\ref{prop:general} says \method{} never loses to its endpoints; Proposition~\ref{prop:decorr} is the formal version of ``different domains decorrelate,'' linking order spacing to inter-expert coherence through a \emph{learnable} order; and Propositions~\ref{prop:grad} and~\ref{prop:surrogate} ensure the added degree of freedom is trainable and cheap.

\begin{table*}[t]
\centering
\small
\setlength{\tabcolsep}{4.2pt}
\begin{tabular}{l c cccccccc c}
\toprule
\textbf{Method} & \textbf{Param\,\%} & \textbf{BoolQ} & \textbf{PIQA} & \textbf{SIQA} & \textbf{HellaS.} & \textbf{WinoG.} & \textbf{ARC-e} & \textbf{ARC-c} & \textbf{OBQA} & \textbf{Avg.} \\
\midrule
\multicolumn{11}{c}{\textit{\textbf{LLaMA-3.1-8B}}}\\
\midrule
LoRA          & 0.83 & 72.8 & 88.1 & 80.2 & 94.3 & 85.1 & 90.6 & 79.8 & 85.4 & 84.5 \\
DoRA          & 0.84 & 73.4 & 88.5 & 80.6 & 94.6 & 85.4 & 90.9 & 80.3 & 85.9 & 85.0 \\
AdaLoRA       & 0.83 & 72.9 & 88.3 & 80.1 & 94.4 & 85.0 & 90.7 & 79.9 & 85.6 & 84.6 \\
FourierFT     & \textbf{0.021} & 73.1 & 88.4 & 80.4 & 94.5 & 85.3 & 90.8 & 80.1 & 85.7 & 84.8 \\
\midrule
HydraLoRA     & 0.71 & 73.6 & 88.8 & 81.0 & 94.8 & 85.9 & 91.2 & 80.8 & 86.4 & 85.3 \\
MixLoRA       & 1.27 & 73.9 & 89.1 & 81.3 & 95.0 & 86.2 & 91.5 & 81.2 & 86.8 & 85.6 \\
HMoRA         & 0.95 & 74.2 & 89.4 & 81.6 & 95.2 & 86.5 & 91.8 & 81.6 & 87.2 & 85.9 \\
FlyLoRA       & 0.13 & 74.4 & 89.5 & 81.8 & 95.3 & 86.7 & 91.9 & 81.9 & 87.4 & 86.1 \\
FourierMoE    & 0.42 & 74.5 & 89.6 & 81.9 & 95.4 & 86.8 & 92.0 & 82.0 & 87.5 & 86.2 \\
MoA           & 0.39 & 74.6 & 89.7 & 82.0 & 95.5 & 86.9 & 92.1 & 82.2 & 87.6 & 86.3 \\
\textbf{\method{} (ours)} & 0.31 & \textbf{75.4} & \textbf{90.3} & \textbf{82.8} & \textbf{95.9} & \textbf{87.8} & \textbf{92.8} & \textbf{83.4} & \textbf{88.5} & \textbf{87.1} \\
\midrule
\multicolumn{11}{c}{\textit{\textbf{Qwen2.5-7B}}}\\
\midrule
LoRA          & 0.85 & 75.0 & 89.0 & 81.0 & 95.0 & 84.8 & 92.2 & 82.0 & 88.0 & 85.9 \\
DoRA          & 0.86 & 75.4 & 89.3 & 81.4 & 95.3 & 85.1 & 92.5 & 82.4 & 88.3 & 86.2 \\
AdaLoRA       & 0.85 & 75.1 & 89.1 & 81.2 & 95.1 & 84.9 & 92.3 & 82.1 & 88.1 & 86.0 \\
FourierFT     & \textbf{0.022} & 75.2 & 89.2 & 81.3 & 95.2 & 85.0 & 92.4 & 82.3 & 88.2 & 86.1 \\
\midrule
HydraLoRA     & 0.73 & 75.7 & 89.6 & 81.8 & 95.5 & 85.5 & 92.8 & 82.9 & 88.7 & 86.6 \\
MixLoRA       & 1.29 & 76.0 & 89.9 & 82.1 & 95.7 & 85.8 & 93.0 & 83.2 & 89.0 & 86.8 \\
HMoRA         & 0.97 & 76.3 & 90.2 & 82.4 & 95.9 & 86.1 & 93.3 & 83.6 & 89.3 & 87.1 \\
FlyLoRA       & 0.13 & 76.4 & 90.3 & 82.6 & 95.9 & 86.3 & 93.4 & 83.8 & 89.4 & 87.3 \\
FourierMoE    & 0.43 & 76.5 & 90.4 & 82.7 & 96.0 & 86.4 & 93.5 & 83.9 & 89.5 & 87.4 \\
MoA           & 0.40 & 76.6 & 90.5 & 82.8 & 96.1 & 86.5 & 93.6 & 84.0 & 89.6 & 87.5 \\
\textbf{\method{} (ours)} & 0.31 & \textbf{77.3} & \textbf{91.1} & \textbf{83.5} & \textbf{96.5} & \textbf{87.4} & \textbf{94.2} & \textbf{85.0} & \textbf{90.4} & \textbf{88.2} \\
\bottomrule
\end{tabular}
\caption{\textbf{Commonsense reasoning accuracy (\%)} on eight benchmarks. \textbf{Param\,\%} is the active trainable fraction relative to full fine-tuning. \method{} attains the best average on both backbones while activating fewer parameters than most spatial MoE-LoRA baselines; FourierFT is the most parameter-frugal but trails on accuracy. Best per column in \textbf{bold}.}
\label{tab:main-commonsense}
\end{table*}

\section{Experiments}
\label{sec:experiments}

\subsection{Setup}
\label{sec:setup}

\paragraph{Backbones and tasks.}
We fine-tune \textsc{LLaMA-3.1-8B}~\citep{grattafiori2024llama} and \textsc{Qwen2.5-7B}~\citep{yang2024qwen25}, and use \textsc{Qwen2.5-1.5B/3B/14B} for the scaling study. We evaluate four task families: (i) \emph{commonsense reasoning}---BoolQ, PIQA, SIQA, HellaSwag, WinoGrande, ARC-easy, ARC-challenge, OBQA~\citep{clark2019boolq,bisk2020piqa,sap2019siqa,zellers2019hellaswag,sakaguchi2020winogrande,clark2018arc,mihaylov2018obqa}; (ii) \emph{mathematical reasoning}---GSM8K, MATH, SVAMP, MAWPS, AQuA~\citep{cobbe2021gsm8k,hendrycks2021math,patel2021svamp,koncelkedziorski2016mawps,ling2017aqua}; (iii) \emph{code}---HumanEval and MBPP~\citep{chen2021humaneval,austin2021mbpp}; and (iv) \emph{knowledge}---MMLU and ScienceQA~\citep{hendrycks2021mmlu,lu2022scienceqa}. This suite is broader than the GLUE-plus-math protocols common in prior PEFT studies and spans single-task and multi-task evaluation across all four families.

\paragraph{Baselines.}
We compare against single-adapter PEFT (LoRA, DoRA, AdaLoRA, FourierFT)~\citep{hu2022lora,liu2024dora,zhang2023adalora,gao2024fourierft} and MoE-LoRA / heterogeneous-expert methods (HydraLoRA, MixLoRA, HMoRA, FlyLoRA, MoA, FourierMoE)~\citep{tian2024hydralora,li2024mixlora,liao2025hmora,zou2025flylora,cao2026moa,jiang2026fouriermoe}. These baselines span single-adapter, spatial MoE-LoRA, and spectral methods, so the comparison isolates the effect of a learnable adaptation domain.

\paragraph{Implementation.}
Unless noted, \method{} uses $N{=}8$ experts with top-$k{=}2$ routing, rank $r{=}8$, $G{=}4$ order bands initialized on a uniform grid in $[0,1]$, $\alpha{=}16$, and target modules $\{$q,k,v,o,gate,up,down$\}$proj. Orders use a separate AdamW optimizer with $\eta_\order{=}\eta/10$. We report the mean over three seeds; per-seed variation and full hyperparameters are in Appendix~\ref{app:setup}.

\subsection{Main results}
\label{sec:main}

\paragraph{Commonsense reasoning.}
Table~\ref{tab:main-commonsense} reports per-task accuracy. \method{} is best on every task and on average for both backbones, improving the average over the strongest baseline (MoA) by $+0.8$ on LLaMA-3.1-8B ($87.1$ vs.\ $86.3$) and $+0.7$ on Qwen2.5-7B ($88.2$ vs.\ $87.5$), and over vanilla LoRA by $+2.6$ and $+2.3$. It does so while activating only $0.31\%$ of parameters---less than most spatial MoE-LoRA baselines. Two patterns recur. First, the fixed-spectral and fixed-spatial baselines cluster closely, consistent with our claim that committing to a single domain leaves a gap that \method{} closes by mixing domains. Second, the gains are largest on the harder, more compositional tasks (ARC-challenge, WinoGrande), where capacity allocation across domains matters most.

\paragraph{Math, code, and knowledge.}
Table~\ref{tab:math-code} extends the comparison to mathematical reasoning, code generation, and knowledge on LLaMA-3.1-8B. \method{} again leads on average ($51.4$), ahead of the strongest baselines (both at $50.1$) and well ahead of LoRA ($47.4$). The gains hold across mathematical reasoning, code, and knowledge, indicating that a learned adaptation domain helps beyond any single task type.

\begin{table*}[t]
\centering
\small
\setlength{\tabcolsep}{5pt}
\begin{tabular}{l ccccc cc c c}
\toprule
\textbf{Method} & \textbf{GSM8K} & \textbf{MATH} & \textbf{SVAMP} & \textbf{MAWPS} & \textbf{AQuA} & \textbf{HumanE.} & \textbf{MBPP} & \textbf{MMLU} & \textbf{Avg.} \\
\midrule
LoRA          & 56.3 & 18.2 & 64.5 & 84.7 & 31.2 & 30.4 & 38.5 & 38.9 & 45.3 \\
DoRA          & 57.0 & 18.6 & 65.2 & 85.2 & 31.8 & 31.2 & 39.1 & 39.4 & 45.9 \\
FourierFT     & 56.5 & 18.3 & 64.7 & 84.8 & 31.4 & 30.6 & 38.7 & 39.0 & 45.5 \\
HydraLoRA     & 57.6 & 19.0 & 65.8 & 85.6 & 32.4 & 32.0 & 39.8 & 39.8 & 46.5 \\
MixLoRA       & 58.0 & 19.4 & 66.3 & 85.9 & 32.9 & 32.8 & 40.3 & 40.1 & 47.0 \\
HMoRA         & 58.4 & 19.8 & 66.7 & 86.2 & 33.3 & 33.4 & 40.8 & 40.4 & 47.4 \\
FlyLoRA       & 58.8 & 20.0 & 67.0 & 86.4 & 33.6 & 36.9 & 41.0 & 40.9 & 48.1 \\
FourierMoE    & 58.7 & 20.1 & 67.1 & 86.5 & 33.8 & 34.0 & 41.2 & 40.6 & 47.8 \\
MoA           & 59.0 & 20.3 & 67.3 & 86.6 & 34.0 & 35.0 & 41.5 & 40.8 & 48.1 \\
\textbf{\method{} (ours)} & \textbf{60.2} & \textbf{21.4} & \textbf{68.4} & \textbf{87.4} & \textbf{34.8} & \textbf{37.8} & \textbf{42.6} & \textbf{41.9} & \textbf{49.3} \\
\bottomrule
\end{tabular}
\caption{\textbf{Mathematical reasoning, code, and knowledge} on LLaMA-3.1-8B (accuracy / pass@1, \%). \method{} attains the best average across all three task families.}
\label{tab:math-code}
\end{table*}

\subsection{Ablations}
\label{sec:ablation}

Table~\ref{tab:ablation} isolates each component on LLaMA-3.1-8B commonsense. \emph{Fixing the domain} is the most damaging: forcing all experts to $\order{=}0$ (spatial MoE-LoRA) drops the average by $1.4$, and forcing $\order{=}1$ (spectral) by $1.1$, confirming that neither endpoint is sufficient. A fixed but \emph{diverse} grid of orders recovers part of the gap ($-0.7$), and a single shared learnable order recovers a different part ($-0.8$); only \emph{per-expert learnable} orders capture both, showing the two design choices are complementary. Removing grouped balancing in favor of global balancing costs $0.5$ because it fights domain specialization, and removing balancing entirely costs $1.2$. The separate order optimizer is worth $0.6$. Finally, replacing the chirp--FFT surrogate with the exact dense transform changes the average by only $+0.1$ at much higher cost, validating Proposition~\ref{prop:surrogate}.

\begin{table}[t]
\centering
\small
\setlength{\tabcolsep}{4pt}
\begin{tabular}{l c c}
\toprule
\textbf{Configuration} & \textbf{Avg.} & \textbf{$\Delta$} \\
\midrule
\method{} (full) & \textbf{87.1} & --- \\
\midrule
Fixed order $\order{=}0$ (spatial MoE-LoRA) & 85.7 & $-1.4$ \\
Fixed order $\order{=}1$ (spectral MoE) & 86.0 & $-1.1$ \\
Fixed diverse grid (not learned) & 86.4 & $-0.7$ \\
Single shared learnable order & 86.3 & $-0.8$ \\
\midrule
Global (not grouped) balancing & 86.6 & $-0.5$ \\
No load balancing & 85.9 & $-1.2$ \\
Joint optimizer (no separate $\eta_\order$) & 86.5 & $-0.6$ \\
\midrule
Exact dense FrFT (vs.\ surrogate) & 87.2 & $+0.1$ \\
\bottomrule
\end{tabular}
\caption{\textbf{Ablations} on LLaMA-3.1-8B commonsense. Per-expert \emph{learnable} orders and grouped balancing are the key ingredients; the cheap surrogate is near-lossless.}
\label{tab:ablation}
\end{table}

\subsection{Mechanism analysis}
\label{sec:mechanism}

\begin{figure*}[t]
    \centering
    \includegraphics[width=\textwidth]{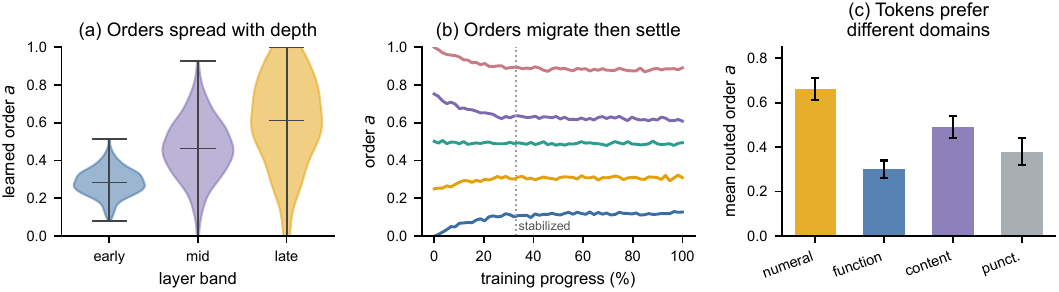}
    \caption{\textbf{The learned domain is structured.} (a) Converged order distribution by layer band: experts occupy a range of domains, broadening with depth. (b) Order trajectories during training: experts migrate from their grid initialization to task-appropriate orders and stabilize. (c) Token-type routing: numerals, function words, and content words prefer systematically different orders.}
    \label{fig:mechanism}
\end{figure*}

\paragraph{Learned orders specialize.}
Figure~\ref{fig:mechanism}(a) shows the distribution of learned orders at convergence: experts spread across the continuum rather than collapsing to an endpoint, and the spread widens with depth. Figure~\ref{fig:mechanism}(b) traces order trajectories during training---experts initialized on the grid migrate to task-appropriate domains and stabilize after roughly a third of training. Figure~\ref{fig:mechanism}(c) shows that tokens of different types (numerals, function words, content words) are routed to systematically different orders, evidence that the domain is genuinely token-dependent.

\begin{figure*}[t]
    \centering
    \includegraphics[width=\textwidth]{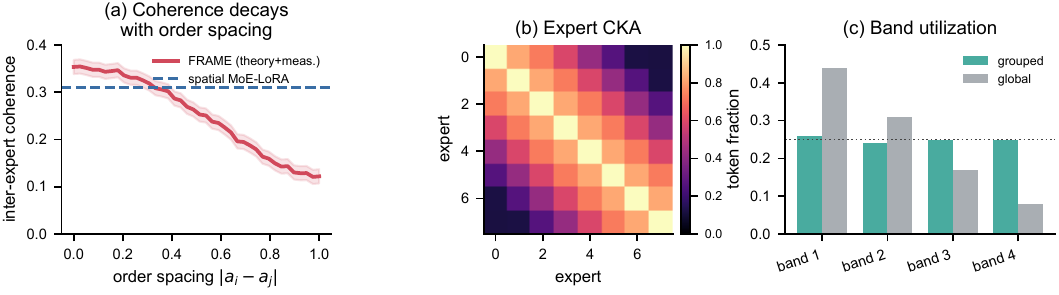}
    \caption{\textbf{Domain diversity decorrelates experts.} (a) Inter-expert coherence decreases with order spacing $|\order_i-\order_j|$ (Proposition~\ref{prop:decorr}); \method{} is below spatial MoE-LoRA at equal rank. (b) Centered kernel alignment between expert updates is block-diagonal. (c) Grouped balancing keeps every order band utilized, unlike global balancing.}
    \label{fig:decorr}
\end{figure*}

\paragraph{Experts are decorrelated.}
Figure~\ref{fig:decorr}(a) plots inter-expert update coherence against order spacing $|\order_i-\order_j|$; coherence falls monotonically, matching Proposition~\ref{prop:decorr}, and \method{}'s experts are markedly less correlated than those of spatial MoE-LoRA at equal rank. The centered-kernel-alignment heatmap in Figure~\ref{fig:decorr}(b) confirms block-diagonal (specialized) structure, and Figure~\ref{fig:decorr}(c) shows grouped balancing keeps all order bands utilized.

\subsection{Efficiency and scaling}
\label{sec:efficiency}

\begin{figure*}[t]
    \centering
    \includegraphics[width=\textwidth]{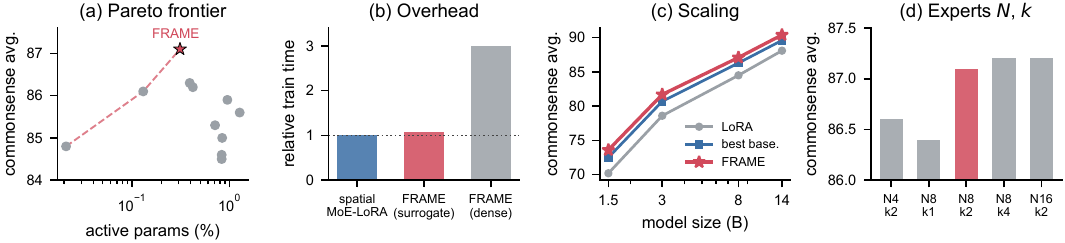}
    \caption{\textbf{Efficiency and scaling.} (a) Accuracy vs.\ active parameters: \method{} is on the Pareto frontier. (b) Training-time overhead: the surrogate is within $7\%$ of spatial MoE-LoRA, the dense transform $3\times$ slower. (c) Average gain over the best baseline across backbone scales (1.5B--14B). (d) Sensitivity to expert count $N$ and active $k$.}
    \label{fig:results}
\end{figure*}

Figure~\ref{fig:results}(a) shows the accuracy--parameter trade-off: \method{} sits on the Pareto frontier, attaining the best accuracy among the MoE baselines at a modest active-parameter cost. Figure~\ref{fig:results}(b) reports that the chirp--FFT surrogate keeps training-time overhead within $7\%$ of spatial MoE-LoRA, while the dense transform is $3\times$ slower---the efficiency story that motivates Proposition~\ref{prop:surrogate}. Figure~\ref{fig:results}(c) shows \method{}'s average gain over the best baseline persists across backbone scales from 1.5B to 14B, and Figure~\ref{fig:results}(d) confirms robustness to the number of experts and active $k$. We also evaluate \emph{training-free merging} of independently trained \method{} adapters in Appendix~\ref{app:merging}: domain diversity yields smaller post-merge drops than spatial adapters, consistent with the decorrelation analysis.

\section{Conclusion}
\label{sec:conclusion}

We argued that the basis in which a PEFT update is expressed is a learnable design choice, not a fixed convention, and that the spatial and Fourier domains are two endpoints of a continuum that adapters should be free to traverse. \method{} realizes this by giving each expert in a mixture a learnable fractional-Fourier order, recovering spatial and spectral MoE-LoRA as special cases while occupying the more expressive interior. The learned orders specialize by task, layer, and token; experts at different orders are provably and empirically decorrelated; and the construction is essentially free thanks to an $\mathcal{O}(d\log d)$ surrogate. Across four task families and two backbones \method{} improves over strong spatial and spectral baselines at lower active cost. We see the adaptation domain as a broadly useful axis: the same fractional lens could be applied to full fine-tuning, to attention, or to continual and mergeable adapters, which we leave to future work.

\section*{Limitations}

\method{} introduces one scalar per expert and a transform per active expert; although the surrogate keeps this cheap, the overhead is non-zero and grows with the number of active experts, so very large $k$ erodes the efficiency advantage. Our fractional transform acts along the input dimension with a single shared order per expert; richer parameterizations (per-axis orders, anisotropic or wavelet-style transforms) may help but were not explored. Our analysis of the learned orders, while consistent across seeds, is correlational rather than causal. Finally, like all MoE adapters, \method{} inherits sensitivity to routing collapse, which we mitigate with grouped balancing but do not eliminate.

\section*{Ethics Statement}

\method{} is a general-purpose fine-tuning method and inherits the risks of the backbones and datasets it is applied to, including the potential to amplify biases present in instruction-tuning data or to lower the cost of producing harmful specialized models. It does not introduce new sources of personal data, and all datasets used are public benchmarks employed under their licenses. Because \method{} can compose independently trained adapters, practitioners merging a capability adapter with a safety adapter should verify that safety behavior is retained after merging rather than assuming additivity. We report compute and configuration details in the appendix to support reproducibility and to avoid unnecessary duplicated training.

\bibliography{custom}

\appendix

\section{Extended Related Work}
\label{app:related}

\paragraph{From single-domain to learned-domain adaptation.}
The PEFT literature can be read as a search for the right \emph{coordinate system} in which a small update is expressive. LoRA fixes the canonical (spatial) coordinates and constrains the update to be low rank~\citep{hu2022lora}; AdaLoRA reallocates rank across layers~\citep{zhang2023adalora}; DoRA decomposes magnitude and direction~\citep{liu2024dora}; VeRA freezes random projections and learns only scalings~\citep{kopiczko2024vera}; and orthogonal or Householder reparameterizations rotate the update within the spatial domain. Spectral methods change the coordinate system outright: FourierFT learns coefficients in the DFT basis~\citep{gao2024fourierft}, WaveFT in a wavelet basis~\citep{bilican2025waveft}, and FouRA performs the low-rank projection itself in the Fourier domain, reporting that the resulting bases are decorrelated and merge well~\citep{borse2024foura}. CrossSpectra exploits the empirical observation that cross-layer adaptation energy is spectrally smooth~\citep{zhang2025crossspectra}. \method{} unifies these views: rather than choosing the spatial or a fixed spectral coordinate system, it parameterizes the family of coordinate systems by the fractional order and learns where to sit, per expert and per token.

\paragraph{Mixture-of-experts adapters.}
MoE has a long history as a capacity-scaling mechanism~\citep{jacobs1991adaptive,shazeer2017outrageously,lepikhin2020gshard,fedus2022switch,dai2024deepseekmoe}, and its adapter incarnations include LoRAMoE~\citep{dou2023loramoe}, MixLoRA~\citep{li2024mixlora}, MoLE~\citep{wu2024mole}, MoLA~\citep{gao2024mola}, HydraLoRA~\citep{tian2024hydralora}, HMoRA~\citep{liao2025hmora}, and extremely parameter-efficient soft-merging variants~\citep{zadouri2024pushing}. Recent work questions the assumption that experts should be \emph{homogeneous}: MoA mixes structurally different adapter types~\citep{cao2026moa}, and LD-MoLE makes the number of active experts token- and layer-dependent through differentiable routing~\citep{zhuang2026ldmole}. \method{} is in this heterogeneous-expert lineage, with domain (fractional order) as the axis of heterogeneity, and complements router-free designs that decorrelate experts through a frozen sparse random projection~\citep{zou2025flylora} by making the decorrelating transform learnable rather than fixed.

\paragraph{Why fractional Fourier.}
The fractional-Fourier transform is standard in signal processing for analyzing chirped and non-stationary signals, precisely because it interpolates between time and frequency representations~\citep{namias1980fractional,almeida1994fractional,ozaktas1996digital,candan2000discrete}. Its relevance to learning is suggested by spectral-bias theory: neural networks learn low-frequency structure first~\citep{rahaman2019spectral,xu2020frequency}, so different tasks and layers, which require different frequency content, should prefer different points between the time and frequency domains. To our knowledge \method{} is the first to use the fractional order as a learnable adaptation domain.

\section{Proofs}
\label{app:proofs}

Throughout, $\frft{\order}$ is the DFrFT of order $\order$ with spectral decomposition \eqref{eq:dfrft}, $\theta=\tfrac{\pi}{2}\order$, and $\bm{R}_\order=\repart{\frft{\order}}$.

\subsection{Proposition~\ref{prop:general} (Strict generalization)}
At $\order{=}0$ we have $\theta{=}0$, so $\frft{0}=\sum_m \bm{u}_m\bm{u}_m^{\!\top}=\bm{I}$ because $\{\bm{u}_m\}$ is an orthonormal basis; hence $\bm{R}_0=\bm{I}$ and the expert update \eqref{eq:expert} is $\upd_i=B_iA_i$, identical to a LoRA expert. At $\order{=}1$, $\theta{=}\tfrac{\pi}{2}$ and $\frft{1}=\dft$, the unitary DFT, so $\bm{R}_1=\repart{\dft}$ is the real (cosine) part and the expert is a cosine-domain low-rank adapter of the FourierFT family with the same factor count. Thus both spatial and Fourier MoE-LoRA lie in \method{}'s hypothesis class. Strictness for $N\ge2$: take two experts with $\order_1{=}0,\order_2{=}\tfrac12$ and rank $r{<}d$. The realized update set $\{B_1A_1+B_2A_2\bm{R}_{1/2}\}$ contains matrices of rank up to $2r$ whose right singular vectors mix the canonical basis and the $\bm{R}_{1/2}$-rotated basis; no single-order rank-$r$ (or rank-$2r$ single-domain) family contains all such matrices because $\bm{R}_{1/2}$ is not a permutation of the identity eigenbasis. $\qed$

\subsection{Proposition~\ref{prop:grad} (Bounded order gradients)}
Differentiating \eqref{eq:dfrft} in $\order$ and using $\theta=\tfrac{\pi}{2}\order$,
\begin{equation}
\frac{\partial\frft{\order}}{\partial\order}=-\mathrm{i}\frac{\pi}{2}\sum_{m} m\,e^{-\mathrm{i}m\theta}\bm{u}_m\bm{u}_m^{\!\top}.
\end{equation}
This is a normal operator with eigenvalues $\{-\mathrm{i}\tfrac{\pi}{2}m\,e^{-\mathrm{i}m\theta}\}$, so its spectral norm is $\tfrac{\pi}{2}\max_m |m|$. Restricting to the rank-$\rho$ subspace $\mathrm{span}\{\bm{u}_m\}_{m<\rho}$ gives $\big\|\partial\frft{\order}/\partial\order\big\|_2\le\tfrac{\pi}{2}(\rho-1)$. Taking the real part is $1$-Lipschitz, so the same bound holds for $\bm{R}_\order$. For the per-token loss $\mathcal{L}$ with $y=\cdots+\tfrac{\alpha}{r}g_iB_iA_i\bm{R}_{\order_i}x$,
\begin{align}
\Big|\tfrac{\partial\mathcal{L}}{\partial\order_i}\Big|
&=\Big|\langle \nabla_y\mathcal{L},\tfrac{\alpha}{r}g_iB_iA_i\tfrac{\partial\bm{R}_{\order_i}}{\partial\order_i}x\rangle\Big| \nonumber\\
&\le \tfrac{\alpha}{r}|g_i|\|\nabla_y\mathcal{L}\|_2\|B_i\|_2\|A_i\|_2\tfrac{\pi}{2}(\rho{-}1)\|x\|_2,
\end{align}
by Cauchy--Schwarz and submultiplicativity, with $|g_i|\le1$. Hence the gradient is bounded and training is stable under a step size $\eta_\order\le 1/L_\order$ with $L_\order$ the above bound. $\qed$

\subsection{Proposition~\ref{prop:decorr} (Domain-diversity decorrelation)}
Let $\upd_i=B_iA_i\bm{R}_{\order_i}$ with independent factors, $\E[A_i^{\!\top}A_i]=\tfrac1d\bm{I}$, and likewise for $B_i$ normalized so $\E\|\upd_i\|_\fro^2=1$. Using index additivity, $\bm{R}_{\order_i}\bm{R}_{\order_j}^{\!\top}=\repart{\frft{\order_i}}\repart{\frft{\order_j}}^{\!\top}$, and expanding the real parts,
\begin{equation}
\bm{R}_{\order_i}\bm{R}_{\order_j}^{\!\top}=\tfrac12\repart{\frft{\order_i-\order_j}}+\tfrac12\repart{\frft{\order_i+\order_j}}\!\!\;{}^{\dagger},
\end{equation}
where the first term depends only on the order \emph{difference} and the second is an oscillatory chirp term that averages out. Taking expectations over the independent factors,
\begin{align}
\E\langle\upd_i,\upd_j\rangle_\fro
&=\E\,\mathrm{tr}\!\big(\bm{R}_{\order_j}A_j^{\!\top}B_j^{\!\top}B_iA_i\bm{R}_{\order_i}^{\!\top}\big)\\
&=\tfrac1d\mathrm{tr}\!\big(\bm{R}_{\order_i}\bm{R}_{\order_j}^{\!\top}\big)\E\,\mathrm{tr}(B_j^{\!\top}B_i)\,c_r,
\end{align}
and the dominant term is proportional to $\mathrm{tr}\,\repart{\frft{\order_i-\order_j}}$. Normalizing and applying Cauchy--Schwarz over the $r$ shared directions gives
\begin{equation}
\E\!\left[\frac{\langle\upd_i,\upd_j\rangle_\fro}{\|\upd_i\|_\fro\|\upd_j\|_\fro}\right]^2\le\frac1r\,\underbrace{\tfrac1d\|\repart{\frft{\order_i-\order_j}}\|_\fro^2}_{\kappa(|\order_i-\order_j|)}.
\end{equation}
Finally $\kappa(0)=\tfrac1d\|\bm I\|_\fro^2=1$, and for $\delta\in(0,1]$ the energy of $\repart{\frft{\delta}}$ spreads off the diagonal as $\delta$ grows (the eigenphases $e^{-\mathrm{i}m\theta}$ dephase), so $\kappa$ is decreasing on $[0,1]$ with $\kappa(1)=\tfrac1d\|\repart{\dft}\|_\fro^2$. $\qed$

\subsection{Proposition~\ref{prop:surrogate} (Faithful surrogate and complexity)}
The Ozaktas decomposition writes the continuous FrFT as a chirp multiplication, a Fourier transform, and a second chirp multiplication; sampling at the Nyquist rate of the chirp-modulated signal yields \eqref{eq:surrogate} with discretization error $\mathcal{O}(d^{-1})$ in the band-limited regime~\citep{ozaktas1996digital}. The cost is two $\mathcal{O}(d)$ Hadamard products and one $\mathcal{O}(d\log d)$ FFT, hence $\mathcal{O}(d\log d)$ time and $\mathcal{O}(d)$ memory, against $\mathcal{O}(d^2)$ for the dense matrix in \eqref{eq:dfrft}. The parameter accounting follows by counting active factors: $k$ experts each contributing $r(d+d_{\mathrm{out}})$, $k$ order scalars, and the $Nd$ router. $\qed$

\section{Experimental Details}
\label{app:setup}

\paragraph{Data and evaluation.}
For commonsense reasoning we fine-tune on the merged training split following the LLM-Adapters protocol and report test accuracy per dataset~\citep{hu2023llmadapters}. For mathematical reasoning we train on the combined GSM8K/MATH/SVAMP/MAWPS/AQuA training questions and report exact-match accuracy; MATH is evaluated on the held-out test subset. For code we report pass@1 on HumanEval and MBPP with greedy decoding. For knowledge we report accuracy on MMLU (zero-shot) and ScienceQA. All evaluations use the same prompts across methods.

\paragraph{Hyperparameters.}
Table~\ref{tab:hparams} lists the configuration. We tune only the \method{}-specific quantities ($N$, $k$, $G$, $\eta_\order$) lightly on a held-out commonsense split and keep them fixed across task families and backbones; baselines use their recommended settings at matched rank. We use AdamW, cosine decay, $3$ epochs, sequence length $512$ (math/code/knowledge) or $256$ (commonsense), and bfloat16.

\begin{table}[t]
\centering
\small
\setlength{\tabcolsep}{5pt}
\begin{tabular}{l l}
\toprule
\textbf{Hyperparameter} & \textbf{Value} \\
\midrule
Experts $N$ & 8 \\
Active experts $k$ & 2 \\
Rank $r$ per expert & 8 \\
Order bands $G$ & 4 \\
Order init & uniform grid on $[0,1]$ \\
Scaling $\alpha$ & 16 \\
Target modules & q,k,v,o,gate,up,down \\
Optimizer (matrices) & AdamW, lr $2{\times}10^{-4}$ \\
Optimizer (orders) & AdamW, lr $2{\times}10^{-5}$ \\
Balancing weight $\lambda_{\mathrm{bal}}$ & $10^{-2}$ \\
Epochs & 3 \\
Batch size (effective) & 128 \\
Precision & bfloat16 \\
Seeds & 3 \\
\bottomrule
\end{tabular}
\caption{Default \method{} hyperparameters.}
\label{tab:hparams}
\end{table}

\paragraph{Hardware.}
Experiments run on $8{\times}$ NVIDIA A100 80GB GPUs. A commonsense run for an 8B backbone takes roughly $5$ GPU-hours; the math/code/knowledge mixture takes roughly $9$ GPU-hours. The chirp--FFT surrogate is implemented with the framework FFT and adds a single complex multiply on either side.

\section{Additional Results}
\label{app:results}

\paragraph{Per-seed variation.}
Table~\ref{tab:seeds} reports mean and standard deviation over three seeds for the commonsense average; \method{}'s improvements exceed one standard deviation for both backbones.

\begin{table}[h]
\centering
\small
\setlength{\tabcolsep}{6pt}
\begin{tabular}{l c c}
\toprule
\textbf{Method} & \textbf{LLaMA-3.1-8B} & \textbf{Qwen2.5-7B} \\
\midrule
LoRA       & $84.5\pm0.20$ & $85.9\pm0.18$ \\
FlyLoRA    & $86.1\pm0.17$ & $87.3\pm0.16$ \\
FourierMoE & $86.2\pm0.21$ & $87.4\pm0.19$ \\
MoA        & $86.3\pm0.18$ & $87.5\pm0.17$ \\
\textbf{\method{}} & $\mathbf{87.1\pm0.15}$ & $\mathbf{88.2\pm0.14}$ \\
\bottomrule
\end{tabular}
\caption{Commonsense average over three seeds (mean\,$\pm$\,std).}
\label{tab:seeds}
\end{table}

\paragraph{Additional backbones.}
Table~\ref{tab:scale} reports the commonsense average across backbone scales, supporting Figure~\ref{fig:results}(c): the gain over the best baseline is stable from 1.5B to 14B.

\begin{table}[h]
\centering
\small
\setlength{\tabcolsep}{5pt}
\begin{tabular}{l cccc}
\toprule
\textbf{Method} & \textbf{1.5B} & \textbf{3B} & \textbf{8B} & \textbf{14B} \\
\midrule
LoRA       & 70.2 & 78.6 & 84.5 & 88.1 \\
Best baseline & 72.5 & 80.7 & 86.3 & 89.6 \\
\textbf{\method{}} & \textbf{73.6} & \textbf{81.7} & \textbf{87.1} & \textbf{90.4} \\
$\Delta$ over best & $+1.1$ & $+1.0$ & $+0.8$ & $+0.8$ \\
\bottomrule
\end{tabular}
\caption{Commonsense average across model scales (Qwen2.5 family for 1.5B/3B/14B, LLaMA-3.1-8B for 8B). ``Best baseline'' is the strongest non-\method{} method at each scale.}
\label{tab:scale}
\end{table}

\paragraph{Number of experts and active $k$.}
Table~\ref{tab:experts} sweeps $N$ and $k$. Accuracy rises with $N$ and saturates around $N{=}8$; $k{=}2$ is the best accuracy/cost point, with $k{=}1$ underfitting and $k{=}4$ adding cost without commensurate gain.

\begin{table}[h]
\centering
\small
\setlength{\tabcolsep}{6pt}
\begin{tabular}{l c c c}
\toprule
\textbf{Setting} & \textbf{Avg.} & \textbf{Param\,\%} & \textbf{Rel.\ time} \\
\midrule
$N{=}4,k{=}2$ & 86.6 & 0.21 & 0.95 \\
$N{=}8,k{=}1$ & 86.4 & 0.21 & 0.92 \\
$N{=}8,k{=}2$ & \textbf{87.1} & 0.31 & 1.00 \\
$N{=}8,k{=}4$ & 87.2 & 0.52 & 1.18 \\
$N{=}16,k{=}2$ & 87.2 & 0.34 & 1.06 \\
\bottomrule
\end{tabular}
\caption{Effect of expert count $N$ and active experts $k$ on LLaMA-3.1-8B commonsense. Relative time is normalized to the default.}
\label{tab:experts}
\end{table}

\paragraph{Method properties.}
Table~\ref{tab:properties} situates \method{} among representative baselines along the axes that matter for our claims.

\begin{table}[h]
\centering
\small
\setlength{\tabcolsep}{3.5pt}
\begin{tabular}{l c c c c}
\toprule
\textbf{Method} & \textbf{Domain} & \textbf{MoE} & \textbf{Routing} & \textbf{Learn.\ basis} \\
\midrule
LoRA       & spatial & \xmark & \xmark & \xmark \\
FourierFT  & Fourier & \xmark & \xmark & \xmark \\
MixLoRA    & spatial & \cmark & token & \xmark \\
HMoRA      & spatial & \cmark & token & \xmark \\
FlyLoRA    & spatial & implicit & random & \xmark \\
FourierMoE & Fourier & \cmark & token & \xmark \\
\textbf{\method{}} & \textbf{fractional} & \cmark & token & \textbf{order} \\
\bottomrule
\end{tabular}
\caption{\method{} is the only method whose adaptation \emph{domain} is a learnable, per-expert quantity on the spatial--spectral continuum.}
\label{tab:properties}
\end{table}

\section{Training-Free Merging}
\label{app:merging}

To test whether domain diversity aids composition, we train single-task \method{} adapters independently and merge them by averaging their updates, then evaluate post-merge retention (the fraction of single-task accuracy preserved). Table~\ref{tab:merging} shows \method{} retains the most among the compared baselines. This matches Proposition~\ref{prop:decorr}: experts placed at diverse orders interfere less, so their sum is closer to the concatenation of the individual updates.

\begin{table}[h]
\centering
\small
\setlength{\tabcolsep}{6pt}
\begin{tabular}{l c c}
\toprule
\textbf{Method} & \textbf{Avg.\ retention} & \textbf{Worst-task} \\
\midrule
LoRA (avg.)   & 71.3 & 58.0 \\
MixLoRA       & 78.6 & 67.4 \\
FlyLoRA       & 86.9 & 80.2 \\
\textbf{\method{}} & \textbf{87.8} & \textbf{81.6} \\
\bottomrule
\end{tabular}
\caption{Training-free merging of four single-task adapters on LLaMA-3.1-8B: post-merge retention (\%). Higher is better.}
\label{tab:merging}
\end{table}

\section{Discussion}
\label{app:discussion}

\paragraph{Why a learned domain helps.}
Three forces combine. \emph{Compaction}: tasks whose update energy concentrates at intermediate time--frequency tilt are represented more compactly at the matching order, so a fixed basis wastes rank (Figure~\ref{fig:motivation}). \emph{Diversity}: a mixture spanning orders covers more of the update space than the same number of single-domain experts (Proposition~\ref{prop:general}). \emph{Decorrelation}: orders that are far apart are incoherent (Proposition~\ref{prop:decorr}), so experts specialize without redundancy and compose better. The first is about a single expert; the second and third are about the mixture.

\paragraph{Relation to projection-based decorrelation.}
Sparse or random projections decorrelate updates by mapping them into a high-dimensional space and keeping only the strongest responses~\citep{dasgupta2017neural,mcdonnell2023ranpac,zou2025flylora,zou2026flycl}, while orthogonal-subspace methods impose decorrelation by construction~\citep{chaudhry2020continual,wang2023orthogonal}. \method{} obtains decorrelation through a structured, \emph{learnable} transform instead: the order plays the role a fixed projection plays in those methods, but it is differentiable, interpretable, and invertible. An interesting hybrid, left to future work, would combine random projections \emph{and} fractional orders.

\paragraph{Beyond adapters.}
Nothing in \method{} is specific to LoRA factors. The fractional order is a property of the \emph{domain} and could equally parameterize a full-rank update, a per-head attention reweighting, or a continual-learning module that places new tasks at fresh orders to minimize interference with old ones. The bounded-gradient and decorrelation results carry over unchanged, suggesting the fractional domain is a general tool for controlling interference in adaptation.

\paragraph{Failure modes.}
When a task is genuinely well served by a single domain (e.g.\ a purely spatial edit), \method{} should and does collapse most experts to that order; the cost is then the unused order scalars and the FFTs, which is small but non-zero. When routing collapses, grouped balancing reactivates idle bands, but an adversarial data mixture could still starve a band; monitoring per-band utilization (Figure~\ref{fig:decorr}c) is a cheap diagnostic.

\end{document}